\documentclass[letterpaper,journal]{IEEEtran}  

\IEEEoverridecommandlockouts                              

\usepackage[utf8]{inputenc}
\usepackage[T1]{fontenc}
\usepackage{graphicx, tipa}
\usepackage{amsmath,amssymb,amsfonts}
\usepackage[caption=false]{subfig}
\usepackage[ruled,linesnumbered, noend]{algorithm2e}
\usepackage{dirtytalk}
\usepackage{hyperref}
\usepackage{gensymb}
\usepackage{tabularx}
\usepackage{multirow}
\usepackage[short]{optidef}
\usepackage{booktabs}
\usepackage{siunitx}
\usepackage{titlesec}
\usepackage{xcolor}
\newcommand{\rev}[1]{\textcolor{black}{#1}}            
\newif\ifanon
\anonfalse 

\title{\LARGE \bf
DeRep: Learning Depth Representations for End-to-End Navigation in Cluttered Environments
}

\ifanon
\author{\rev{Anonymous Authors}}
\else
\author{Hanyu Jin$^{1}$
and Kenji Shimada$^{1}$
\thanks{$^{1}$The authors are with the Department of Mechanical
Engineering, Carnegie Mellon University, Pittsburgh, PA 15213, USA.
{\tt\small hanyujin@andrew.cmu.edu}}%
}
\fi

\begin{document}

\maketitle
\thispagestyle{empty}
\pagestyle{empty}

\begin{abstract}
End-to-end navigation in cluttered environments relies
on compact visual representations to support collision avoidance and goal-directed motion. Existing methods typically learn these
representations jointly with the policy or through observation reconstruction, which does not explicitly specify
what information the representation should preserve. Moreover, they are usually
evaluated with a single downstream policy learner, although the
benefit of a representation may depend on how the policy is
learned. It therefore remains unclear how different
representation-learning objectives affect downstream navigation
performance and robustness.
This work proposes DeRep, a navigation-oriented representation-learning method that combines local bird’s-eye-view reachability supervision with depth reconstruction. The joint objective encourages the encoder to capture local free-space connectivity while retaining scene information beyond the reachability targets. To compare representation objectives under controlled conditions, we pretrain encoders with a common architecture and freeze them during downstream learning. We evaluate the representations using a linear policy probe, flow-matching imitation learning, and reinforcement learning with PPO. DeRep achieves the lowest linear policy probe prediction error and outperforms
reconstruction alone across all six tested PPO conditions
by 15.1-21.3 percentage points in mean success rate,
achieving the highest mean success rate in four conditions. In imitation learning, both reachability-based objectives achieve higher mean success rates than reconstruction alone, with similar mean success rates between the reachability-only
and joint objectives. These results demonstrate consistent benefits of incorporating
reachability supervision into depth representation pretraining
across different downstream learning methods and evaluation
conditions. We further deploy the PPO policy with the frozen DeRep encoder in Gazebo and on a physical quadrotor without fine-tuning,
demonstrating its capability for zero-shot sim-to-sim and
sim-to-real transfer.

\end{abstract}

\section{INTRODUCTION}
Safe navigation is essential for robots performing tasks such as
inspection and exploration in cluttered environments
\cite{FUEL}\cite{ASIP}.
Modular navigation systems typically separate perception, mapping,
planning, and control \cite{FastPlanner}\cite{EgoPlanner}.
End-to-end approaches instead learn a direct mapping from depth
images and robot state to actions or trajectories
\cite{VitFly}\cite{MAVRL}.
For depth-based policies, an encoder typically compresses the depth
image into a latent representation on which the policy bases its
decisions. Transferring a policy from simulation to the real world
therefore depends in part on whether this representation continues
to provide useful navigation information under conditions not
encountered during training. 

Existing learning-based navigation methods often train the encoder
jointly with the policy \rev{from scratch}, allowing the downstream learning objective
to shape the representation \cite{VitFly}\cite{HighSpeedRL}.
This provides task-driven supervision but does not explicitly specify
which information about the environment the representation should
preserve.
Other methods use reconstruction-based objectives to learn depth
representations, either by reconstructing the original observation
\cite{MAVRL} or a transformed depth image that accounts for the
robot's size \cite{CollisionEnc}.
While reconstruction-based objectives encourage the preservation
of scene geometry, reconstruction performance alone does not
establish how effectively the resulting features support
downstream navigation.

\label{sec:introduction}
\begin{figure}[t]
    \centering
    \includegraphics[width=\linewidth]{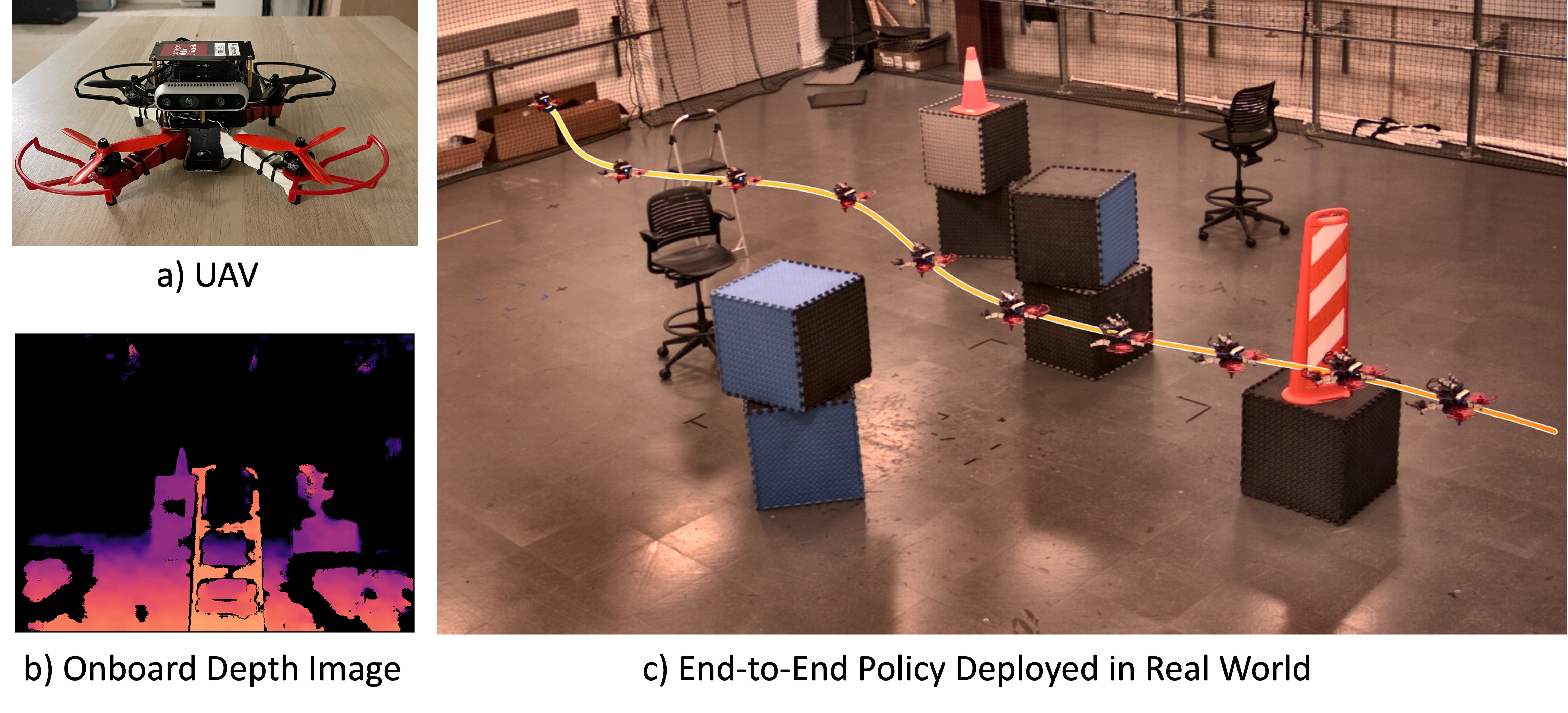}
    \caption{Real-world deployment of DeRep. (a) The quadrotor platform,
equipped with a RealSense D435i depth camera and an onboard Jetson
Orin NX. (b) An onboard depth image. (c) A collision-free flight in the real world.}
    \label{fig:cover}
\end{figure}

In this work, we ask what information a depth representation should
preserve to support collision-free, goal-directed navigation.
Specifically, we investigate whether explicitly
supervising local free-space connectivity provides benefits
beyond depth reconstruction alone.
The usefulness of a representation may also depend on how the
downstream policy is learned and on the evaluation conditions
\cite{pmlr-v202-hu23h}.
A controlled comparison is therefore needed to examine how
representation learning objectives affect navigation performance
and robustness under changes in environment geometry and
depth observations.

To investigate these questions, we propose DeRep, a navigation-oriented representation
learning method that combines local reachability supervision with
depth reconstruction.
The reachability objective predicts an ego-centric bird's-eye-view (BEV)
map of reachable space and the associated \rev{shortest-path distance to each reachable cell}, encouraging
the representation to capture local free-space connectivity (Sec.~\ref{sec:method_reach}).
Depth reconstruction complements this objective by encouraging
the preservation of \rev{pixel-level depth information} (Sec.~\ref{sec:method_joint}).

We evaluate the representations using a linear policy probe, flow-matching imitation learning, and reinforcement learning with PPO. The linear \rev{policy} probe tests whether expert future displacements can be predicted from the frozen representation and navigation state using a linear readout (Sec.~\ref{sec:probe_eval}). The two downstream policy learners then evaluate whether the representations support closed-loop navigation under imitation learning (Sec.~\ref{sec:il_eval}) and reinforcement learning (Sec.~\ref{sec:rl_eval}). \rev{For reinforcement learning, we further evaluate PPO-trained policies under out-of-distribution (OOD) obstacle geometries unseen during training, and under depth corruption including
Gaussian noise, pixel dropout, and quantization, to examine robustness to changes in environment geometry and depth observations.} Finally, we compare the PPO policies qualitatively in Gazebo (Sec.~\ref{sec:gazebo_eval}) and deploy the policy with the DeRep encoder on a physical quadrotor without fine-tuning (Sec.~\ref{sec:real_eval}).

The main contributions of this work are:
\begin{itemize}
    \item A navigation-oriented representation-learning method
    that combines BEV reachability supervision with depth
    reconstruction to learn features for goal-directed navigation.

    \item A systematic evaluation of representation objectives
    using a linear \rev{policy} probe, flow-matching imitation learning,
    and PPO.
    \item Zero-shot sim-to-sim and sim-to-real deployment
    of the PPO policy with the frozen DeRep encoder,
    demonstrated through a qualitative comparison in Gazebo simulations 
    and physical flight experiments.
\end{itemize}

\section{RELATED WORK}

\subsection{Visual Representation Learning}

Visual representation learning aims to encode high-dimensional visual observations into compact features that support downstream
tasks. General-purpose methods such as SimCLR~\cite{SimCLR} and
DINOv2~\cite{dinov2} learn broadly transferable features through
self-supervised objectives. Robotics-oriented methods further adapt
representation learning to embodied data. For example, R3M~\cite{R3M} learns
reusable visual features from egocentric human videos for robot manipulation,
while DeFM~\cite{DeFM} learns large-scale depth representations for transfer
across robotic tasks.

The usefulness of a visual representation, however, can depend
strongly on the information required by the downstream task. Dong
et al.~\cite{VisualControl} show that the ability to decode environment
structure from a representation strongly correlates with downstream control
performance.

Task-relevant representations have also been explored in
embodied navigation, where tasks such as ObjectNav and
PointNav involve different requirements for semantic
understanding, localization, and spatial memory.
Representative works learn navigational features by contrasting egocentric
views against semantic maps~\cite{Ego2Map}, optimizing for navigability through
auxiliary agents~\cite{Mole}, or pretraining on large-scale indoor navigation
datasets~\cite{OVRL}.
Separately, STERLING~\cite{STERLING} learns terrain-cost
representations from unconstrained outdoor robot experience.

These studies highlight the importance of aligning visual
representations with the information required by downstream tasks.

\subsection{End-to-End Navigation}

\rev{End-to-end navigation methods have been widely adopted in cluttered
environments.}
On the reinforcement learning side,
Zhang~et~al.~\cite{HighSpeedRL} equip a depth-based RL policy with a safety
shield for high-speed flight.
On the imitation learning side, Bhattacharya~et~al.~\cite{VitFly} train a
ViT--LSTM architecture via behavior cloning from a privileged expert for
depth-based quadrotor obstacle avoidance.
NoMaD~\cite{NoMaD} adopts a diffusion policy for RGB-based
navigation, while FlowNav~\cite{FlowNav} uses flow matching together with
visual features from a pretrained monocular-depth model.
These methods primarily focus on downstream policy learning and
overall navigation performance rather than isolating how the encoder training
objective affects the learned representation.

A second line of work augments policy learning with reconstruction- or prediction-based representation objectives.
Kulkarni and Alexis~\cite{CollisionEnc} pretrain a depth
encoder to preserve robot-size-aware collision information and use the
resulting latent representation as input to an RL navigation policy.
Kulkarni et al.~\cite{CollisionPrediction} instead learn a semantically
enhanced VAE that preserves thin obstacles and supports collision prediction
over candidate action sequences.
LongRange~\cite{LongRange} combines a pretrained
variational depth encoder with a spatially enhanced recurrent
memory for mapless long-range navigation.
NavRep~\cite{NavRep}, NavDreams~\cite{NavDream}, and MAVRL~\cite{MAVRL}
combine reconstruction with temporal prediction in world-model-style
objectives across LiDAR, RGB, and depth modalities, respectively.

These methods demonstrate the usefulness of learned representations within their respective navigation systems. However, differences in encoder architectures, policy learners, and training settings make it difficult to isolate the effects of representation-learning objectives. In this work, we compare reconstruction, variational regularization, and local reachability supervision using a common encoder architecture and pretraining dataset. The pretrained encoders are frozen
and evaluated using a linear policy probe, flow-matching
imitation learning, and PPO to examine their effects on
navigation performance.

\begin{figure*}[t]
    \centering
    \includegraphics[width=0.96\textwidth]{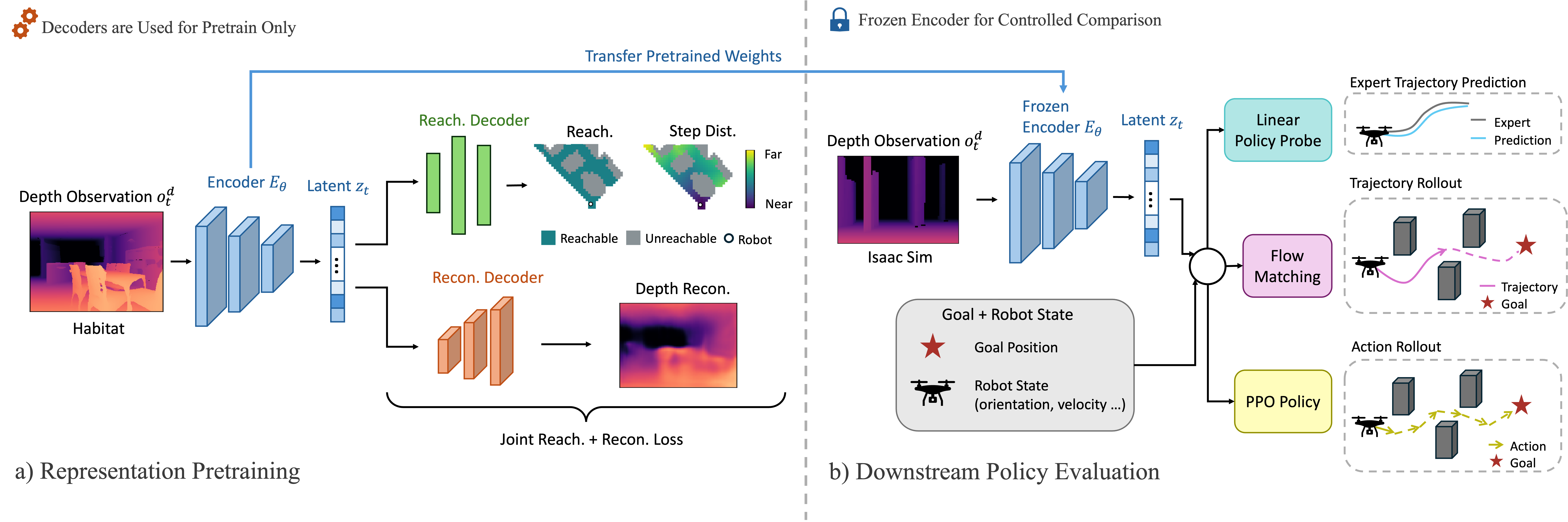}
    \caption{Overview of the proposed framework. (a) Representation
pretraining: the depth encoder is jointly trained with a local
BEV reachability decoder and a depth reconstruction decoder. (b)
Downstream evaluation: after pretraining, the decoders are discarded
and the frozen encoder is evaluated with a linear policy probe,
flow-matching imitation learning, and PPO.
    }
    \label{fig:pretraining_framework}
\end{figure*}

\section{Problem Formulation}
We consider collision avoidance and goal-directed navigation in geometrically cluttered environments. At each time step $t$, the robot receives a depth observation $\mathbf{o}_t^d \in \mathbb{R}^{H \times W}$ and a navigation state $\mathbf{s}_t$, which contains relative goal information and robot state. A depth encoder $E_{\theta}$ maps $\mathbf{o}_t^d$ to a latent representation $\mathbf{z}_t$,
\begin{equation}
    \mathbf{z}_t = E_{\theta}(\mathbf{o}_t^d),
\end{equation}
which is combined with $\mathbf{s}_t$ by a downstream policy $\pi_{\phi}$ to generate a navigation action.

Our objective is to learn an encoder $E_{\theta}$ whose latent representation $\mathbf{z}_t$ effectively supports downstream collision avoidance and goal-directed navigation. We focus on representations extracted from a single depth frame and study how the choice of pretraining objective affects the utility of $\mathbf{z}_t$ for this task.

\section{Methodology}

We pretrain the depth encoder $E_{\theta}$ using a joint objective that combines local BEV reachability prediction and depth reconstruction \rev{as shown in Fig. ~\ref{fig:pretraining_framework} (a).} The two objectives encourage the latent representation $\mathbf{z}_t$ to preserve local free-space connectivity and pixel-level depth information, respectively.

\subsection{Local BEV Reachability}
\label{sec:method_reach}
We define the reachability target on a local ego-centric BEV map
centered at the robot and aligned with its current heading.
The map is constructed based on the ground-truth navigation mesh without additional obstacle inflation,
while the map range is restricted to the current depth camera
field of view. $\Omega_t$ is the set of grid cells
within this local map, with $u$ indexing a grid cell.

Starting from the robot cell, we perform a four-connected flood
fill over the navigable cells in the local BEV map. Each cell is
assigned a binary label $m_{t,u} \in \{0,1\}$ indicating whether
it belongs to the connected navigable component containing the
robot. The set of reachable cells is
$\mathcal{R}_t \subseteq \Omega_t$.
For every reachable cell, we additionally record its shortest-path
distance from the robot, measured in grid steps within the local
map. The step distance is clipped at $d_{\max}$ and normalized
by $d_{\max}$ to obtain the label $c_{t,u}\in[0,1]$. \rev{Examples of the binary reachability and shortest-path distance maps are shown in Fig.~\ref{fig:pretraining_framework} (a).}

\subsection{Joint Representation Learning}
\label{sec:method_joint}

\textbf{Reachability prediction.}
The reachability decoder $D_{\psi}$ predicts two spatially aligned
outputs from $\mathbf{z}_t$: binary reachability logits
$\hat{\mathbf{m}}_t$ and normalized step-distance estimates
$\hat{\mathbf{c}}_t$. Their values at cell $u$ are
$\hat{m}_{t,u}$ and $\hat{c}_{t,u}$, respectively.

The binary reachability loss is a class-weighted cross-entropy
evaluated over $\Omega_t$:
\begin{equation}
\mathcal{L}_{\mathrm{bin}}
=
\frac{1}{|\Omega_t|}
\sum_{u \in \Omega_t}
\operatorname{BCE}_{w}
\left(
\hat{m}_{t,u},m_{t,u}
\right),
\end{equation}
where $\operatorname{BCE}_{w}$ is binary cross-entropy
applied to the predicted logits, with positive-class weight $w$.
The step-distance loss is a Huber regression loss evaluated only
over $\mathcal{R}_t$:
\begin{equation}
\mathcal{L}_{\mathrm{step}}
=
\frac{1}{|\mathcal{R}_t|}
\sum_{u \in \mathcal{R}_t}
\mathcal{H}_{\delta}
\left(
\hat{c}_{t,u}-c_{t,u}
\right),
\end{equation}
where $\mathcal{H}_{\delta}$ is the Huber loss with threshold $\delta$. The two terms are combined using the uncertainty
weighting defined below:
\begin{equation}
\begin{aligned}
\mathcal{L}_{\mathrm{reach}}
={}&
e^{-\lambda_{\mathrm{bin}}}\mathcal{L}_{\mathrm{bin}}
+\lambda_{\mathrm{bin}}
\\
&+
\frac{1}{2}
\left(
e^{-\lambda_{\mathrm{step}}}\mathcal{L}_{\mathrm{step}}
+\lambda_{\mathrm{step}}
\right).
\end{aligned}
\end{equation}
Cells outside $\Omega_t$ contribute to neither loss.
Both outputs are predicted solely from $\mathbf{z}_t$ without
using the navigation state $\mathbf{s}_t$; therefore, the
pretraining objective is independent of a particular navigation
goal.

\textbf{Depth reconstruction.}
To complement the reachability objective, the shared encoder
$E_{\theta}$ and reconstruction decoder $D_{\omega}$ form a
standard depth autoencoder. The decoder reconstructs the input
depth observation from $\mathbf{z}_t$:
\begin{equation}
\hat{\mathbf{o}}_t^d = D_{\omega}(\mathbf{z}_t).
\end{equation}
The reconstruction loss is the mean squared error over valid
depth pixels:
\begin{equation}
\mathcal{L}_{\mathrm{rec}}
=
\frac{1}{|\mathcal{V}_t|}
\sum_{p \in \mathcal{V}_t}
\left(
\hat{o}_{t,p}^d-o_{t,p}^d
\right)^2,
\end{equation}
where $\mathcal{V}_t$ is the set of valid depth pixels,
and $p$ indexes a pixel.
While the reachability target explicitly captures local
navigable-space connectivity and path distance, it abstracts
away fine-grained geometric information in the original
observation. The reconstruction objective therefore encourages
$\mathbf{z}_t$ to retain observation-level geometry that is not
directly supervised by the reachability objective.

\textbf{Combined objective.}
We jointly optimize the shared encoder $E_{\theta}$ and the two
decoders $D_{\psi}$ and $D_{\omega}$ using the binary
reachability, step-distance, and depth-reconstruction losses.
Because these classification and regression objectives have
different numerical scales, we balance their contributions
using homoscedastic uncertainty weighting~\cite{kendall2018}.
Specifically, each loss is associated with a learnable
log-variance parameter $\lambda_i=\log\sigma_i^2$,
where $i \in \{\mathrm{bin},\mathrm{step},\mathrm{rec}\}$.
The complete pretraining objective is
\begin{equation}
\mathcal{L}_{\mathrm{pre}}
=
\mathcal{L}_{\mathrm{reach}}
+
\frac{1}{2}
\left(
e^{-\lambda_{\mathrm{rec}}}\mathcal{L}_{\mathrm{rec}}
+\lambda_{\mathrm{rec}}
\right).
\end{equation}
The log-variance parameters are optimized jointly with the network parameters to adaptively balance the three losses. 
After pretraining, both decoders are discarded, and only
$E_{\theta}$ is retained for the downstream navigation policy.

\section{Experimental Setup}
\label{sec:experimental_setup}

\subsection{Representation Pretraining}

All pretrained encoders share the same CNN architecture with a latent dimension of 256 and are trained on approximately $1.15$ million
single-frame depth observations rendered from Gibson scenes in
Habitat~\cite{xia2018gibson,savva2019habitat}. Each observation is paired with its corresponding local BEV
reachability target, as described in
Sec.~\ref{sec:method_reach}.
We use Gibson to expose the encoders to diverse reconstructed
indoor geometry that is distinct from the synthetic obstacle
layouts used for downstream policy training.
Depth observations are rendered at $60\times80$, clipped at
$5\,\mathrm{m}$, \rev{consistent with the reliable sensing range of
common onboard depth cameras,} and normalized to $[0,1]$.

We evaluate the following encoder pretraining and initialization
strategies:
\begin{itemize}
    \item \textbf{Scratch}: the encoder is randomly initialized
    and trained jointly with the downstream policy.

    \item \textbf{AE}: deterministic autoencoding using depth
    reconstruction alone.

    \item \textbf{VAE}: variational autoencoding using depth
    reconstruction with KL-divergence regularization.

    \item \textbf{Reach (Ours, reachability-only)}: local BEV reachability prediction
    without depth reconstruction, as described in
    Sec.~\ref{sec:method_reach}, serving as an ablation of DeRep.

    \item \textbf{DeRep (Ours)}:
    the proposed joint pretraining objective described in
    Sec.~\ref{sec:method_joint}.
\end{itemize}
These baselines represent the main paradigms in prior work:
Scratch follows end-to-end training of the encoder with the
policy~\cite{VitFly,HighSpeedRL}, AE corresponds to
reconstruction-based depth encoders~\cite{MAVRL}, and VAE
corresponds to variational depth encoders~\cite{CollisionPrediction,LongRange}.

For all pretrained variants, the encoder parameters remain frozen
throughout downstream policy training, allowing us to compare
the representations without further adaptation under a fixed
policy-learning protocol. Scratch serves as an end-to-end reference, with the encoder and policy trained jointly
from random initialization.

\subsection{Downstream Policy Learning}
\label{sec:downstream_policy}
We evaluate the representations through expert trajectory
prediction and closed-loop navigation \rev{as shown in Fig. ~\ref{fig:pretraining_framework}~(b).} First, a linear policy
probe predicts the future displacements of a privileged expert
from the frozen representation and navigation state.
We then evaluate navigation success and safety using two
downstream policies: a flow-matching policy trained on fixed
expert demonstrations and a PPO policy trained through
reward-driven interaction. The expert trajectory dataset contains around $1{,}500$ episodes collected from rollouts of a NavRL~\cite{NavRL} expert, which is used for both the probe and flow-matching policies.
%
Within each evaluation setting, all representation variants use
the same downstream architecture and training protocol.

\paragraph{Linear policy probe.}
Given the per-frame representation $\mathbf{z}_t$, current
velocity $\mathbf{v}_t$, and goal offset $\mathbf{g}_t$,
the probe predicts
\begin{equation}
\hat{\mathbf{y}}_t
=
W_a
\left(
\mathbf{z}_t
+
W_v\mathbf{v}_t
+
W_g\mathbf{g}_t
\right),
\label{eq:linear_probe}
\end{equation}
where $W_v$ and $W_g$ project the velocity and goal offset to
the latent dimension, and $W_a$ maps the combined feature to
the target.
\rev{These linear mappings are trained without nonlinear activations,
so that the probe reflects how directly navigation-relevant
information can be read out from the frozen representation~\cite{SimCLR}.}
The target $\mathbf{y}_t$ contains the expert’s displacements from the current position at each of the next 16 steps, expressed in the goal frame.
We use future displacement because next-step velocity is largely
predictable from the current velocity alone, making it less
informative for comparing visual representations.
The probe is trained using mean squared error (MSE) and
evaluated on held-out episodes.

\paragraph{Imitation learning policy.}

The flow-matching policy is trained by imitation learning on
expert demonstrations and evaluated in closed-loop rollouts.
The policy is conditioned on a two-step observation history.
The conditioning vector is
$\mathbf{h}_t =
[\mathbf{s}_{t-1},\mathbf{z}_{t-1},
 \mathbf{s}_t,\mathbf{z}_t]$.
 
The policy predicts an action chunk
$\mathbf{A}\in\mathbb{R}^{16\times3}$. Each row of $\mathbf{A}$ represents a position increment
$\mathbf{p}_{t+h+1}-\mathbf{p}_{t+h}$ over a
$0.1\,\mathrm{s}$ control interval, for $h=0,\ldots,15$, where $\mathbf{p}_t$ is the robot position.
The increments are expressed in a goal frame fixed at
episode start, with its $z$-axis aligned with the world
vertical and its $x$-axis pointing along the horizontal
start-to-goal direction.
Let $\mathbf{A}_1$ be the normalized expert action chunk
and $\mathbf{A}_0$ a noise chunk with independent standard
Gaussian entries.
We train the flow network $f_\eta$ using the conditional
flow-matching objective along a linear probability path:

{
\begin{equation}
\begin{aligned}
\mathbf{A}_\tau
&=
(1-\tau)\mathbf{A}_0+\tau\mathbf{A}_1,
\qquad \tau\sim\mathcal{U}[0,1],
\\
\mathcal{L}_{\mathrm{FM}}
&=
\mathbb{E}_{\tau,\mathbf{A}_0,(\mathbf{A}_1,\mathbf{h}_t)}
\Big[
\big\lVert
f_\eta(\mathbf{A}_\tau,\tau,\mathbf{h}_t)
-
(\mathbf{A}_1-\mathbf{A}_0)
\big\rVert_F^2
\Big].
\end{aligned}
\label{eq:cfm}
\end{equation}
}

The network regresses the constant target flow velocity
$\mathbf{A}_1-\mathbf{A}_0$ along the straight path from
noise to data.
At inference, we integrate the learned flow from Gaussian
noise to generate an action chunk and execute the first action before replanning at the next time step.

\paragraph{Reinforcement learning policy.}

We train the RL policies using PPO~\cite{schulman2017ppo}
in Isaac Sim.
The actor and critic use separate output heads on a shared
feature network, and both receive the latent representation
and navigation state $[\mathbf{z}_t,\mathbf{s}_t]$.

The action $\mathbf{a}_t\in\mathbb{R}^3$ is a velocity command
expressed in the goal frame.
The command is rotated into the world frame for execution.
The reward at each step is
\begin{equation}
\begin{aligned}
r_t ={}&
\underbrace{
(\mathbf{v}_t^{\mathrm{w}})^{\top}
\hat{\mathbf{g}}_t}
_{\text{progress rate}}
+
\underbrace{(d_{t-1}-d_t)}_{\text{goal progress}}
+
\underbrace{1}_{\text{alive}}
\\
&+
w_{\mathrm{safe}}r^{\mathrm{safe}}_t
-
w_{\mathrm{sm}}
\lVert\mathbf{v}_t^{\mathrm{w}}
-\mathbf{v}_{t-1}^{\mathrm{w}}\rVert_2
-
w_h p^h_t .
\end{aligned}
\label{eq:reward}
\end{equation}

Here, $\mathbf{v}_t^{\mathrm{w}}$ is the world-frame
velocity, $\hat{\mathbf{g}}_t$ is the unit vector toward the
goal in the world frame, and $d_t$ is the Euclidean
goal distance.
The safety reward
$r^{\mathrm{safe}}_t
=\frac{1}{M}\sum_{m=1}^{M}
\log\big(\mathrm{clip}(\rho_{t,m},\epsilon,R_L)\big)$
is a logarithmic function of the safety-lidar ray
distances $\rho_{t,m}$.
$M$ is the number of rays, and $\epsilon>0$
and $R_L$ are the lower and upper clipping bounds,
respectively. $p^h_t$ is a quadratic penalty outside the per-episode
flight-height band.
The reward weights $w_{\mathrm{safe}}$, $w_{\mathrm{sm}}$,
and $w_h$ are fixed across all representation variants.

\section{RESULTS AND DISCUSSION}


\begin{table}[t]
\centering
\caption{Decoding analysis with frozen encoders}
\label{tab:rep_analysis}
\small
\setlength{\tabcolsep}{5pt}
\renewcommand{\arraystretch}{1.1}
\begin{tabular}{lccc}
\toprule
Method & Recon.\ loss $\downarrow$ & Reach.\ acc. $\uparrow$ & Step err.\ $\downarrow$ \\
\midrule
VAE      & 0.0022 & 0.85 & 0.30 \\
AE       & \textbf{0.0012} & 0.87 & 0.29 \\
Reach    & 0.0043 & \textbf{0.90} & \textbf{0.20} \\
DeRep    & 0.0021 & \textbf{0.90} & 0.21 \\
\bottomrule
\end{tabular}
\end{table}

\begin{table*}[t]
    \centering
    \caption{Representation comparison across linear probing and downstream policy learning}
    \label{tab:representation_comparison_full}
    \footnotesize
    \setlength{\tabcolsep}{2.6pt}
    \renewcommand{\arraystretch}{1.1}
    \begin{tabular}{lccccccccc}
        \toprule
        \multirow{3}{*}{Method} & Linear Policy Probe & \multicolumn{2}{c}{Imitation Learning} & \multicolumn{6}{c}{Reinforcement Learning} \\
        \cmidrule(lr){3-4} \cmidrule(lr){5-10}
        & \multirow{2}{*}{Pred. Err. $\downarrow$} & \multirow{2}{*}{Coll./Ep. $\downarrow$} & \multirow{2}{*}{SR (\%) $\uparrow$} & \multirow{2}{*}{ID SR (\%) $\uparrow$} & \multicolumn{2}{c}{OOD SR (\%) $\uparrow$} & \multicolumn{3}{c}{Noise SR (\%) $\uparrow$} \\
        \cmidrule(lr){6-7} \cmidrule(lr){8-10}
        & & & & & Tilt & Concave & Low & Mid & High \\
        \midrule
        Scratch & --     & 6.46 $\pm$ 0.21          & 0.0 $\pm$ 0.0           & 33.5 $\pm$ 19.1          & 21.0 $\pm$ 0.9           & 0.0 $\pm$ 0.0            & 12.8 $\pm$ 1.8          & 5.1 $\pm$ 1.6           & 2.9 $\pm$ 3.7 \\
        VAE     & 0.0288 & 1.34 $\pm$ 0.18          & 24.5 $\pm$ 4.2          & 43.0 $\pm$ 12.4          & 37.0 $\pm$ 9.9           & 18.8 $\pm$ 4.6           & 43.2 $\pm$ 9.9          & 45.2 $\pm$ 2.5          & 44.8 $\pm$ 1.8 \\
        AE      & 0.0274 & 1.19 $\pm$ 0.06          & 31.2 $\pm$ 3.9          & 67.3 $\pm$ 6.1           & 59.3 $\pm$ 4.5           & 31.2 $\pm$ 16.9          & 65.2 $\pm$ 7.5          & 55.5 $\pm$ 16.5         & 35.0 $\pm$ 7.7 \\
        Reach   & 0.0267 & \textbf{0.92} $\pm$ 0.15 & 37.5 $\pm$ 4.2          & 68.7 $\pm$ 6.5           & 63.3 $\pm$ 0.8           & \textbf{56.2} $\pm$ 15.0 & 68.4 $\pm$ 3.0          & 67.1 $\pm$ 4.3          & \textbf{59.4} $\pm$ 6.7 \\
        DeRep   & \textbf{0.0251} & 1.07 $\pm$ 0.14 & \textbf{37.7} $\pm$ 8.7 & \textbf{84.0} $\pm$ 6.1  & \textbf{77.3} $\pm$ 8.2 & 52.5 $\pm$ 10.9          & \textbf{86.5} $\pm$ 6.2 & \textbf{74.5} $\pm$ 3.8 & 50.1 $\pm$ 11.9 \\
        \bottomrule
    \end{tabular}
\end{table*}

To examine the information captured by each representation, we freeze each pretrained encoder and train new decoders for depth reconstruction and BEV reachability prediction.
As shown in Table~\ref{tab:rep_analysis}, AE achieves the lowest reconstruction loss, while Reach and DeRep achieve higher binary reachability accuracy and lower step-distance error than AE and VAE. Compared with Reach, DeRep reduces
reconstruction loss while maintaining similar reachability
accuracy and step-distance error.
These results indicate that the joint representation supports
decoding both depth and reachability information.
\subsection{Linear Policy Probe Evaluation}
\label{sec:probe_eval}
We use a linear \rev{policy} probe to evaluate whether expert future displacement can be
predicted from the frozen pretrained representations. Scratch is excluded
from this evaluation because it involves joint training of the encoder
and downstream policy.
As shown in Table~\ref{tab:representation_comparison_full}, DeRep achieves the lowest
prediction error, followed by Reach, AE, and VAE. Both reachability-based objectives achieve lower prediction errors
than reconstruction alone, with the joint objective achieving
the lowest error.
\subsection{Imitation Learning Evaluation}
\label{sec:il_eval}
We next evaluate the representations through closed-loop navigation using
the flow-matching policy described in
Sec.~\ref{sec:downstream_policy}. All policies are trained on the same
expert demonstrations. Pretrained encoders are kept frozen, while
Scratch is trained jointly with the policy.
We report
the mean and standard deviation of success rate and collisions
per episode across three training seeds for each objective. Success rate is defined as the proportion of evaluation episodes in which the robot reaches the goal without any collisions. Flow-matching and PPO policies are evaluated using the same
in-distribution environments and start-goal
sampling protocol as shown in Fig.~\ref{fig:eval_conditions}~(a). Obstacles are distributed within a $40\times40\,\mathrm{m}$
area. For evaluation, we construct 200 start-goal pairs
with evenly spaced $x$-coordinates in $[-16,16]\,\mathrm{m}$,
with the starts at $y=-24\,\mathrm{m}$ and the goals at
$y=24\,\mathrm{m}$\rev{, so that each episode crosses the entire obstacle field}.

As shown in Table~\ref{tab:representation_comparison_full}, all pretrained
representations achieve fewer collisions and higher success rates
than Scratch, which fails to reach the goal without collisions
in any evaluated episode. \rev{The absolute success rates remain low for all representations.
We attribute this to two factors. First, the privileged expert has
access to a ground-truth map and a wider field of view, whereas the
student policy relies only on onboard depth observations. Second,
behavior cloning suffers from compounding errors in closed-loop
execution. We therefore focus on the relative comparison of success
rates under this shared imitation setup.}
Both reachability-based objectives achieve lower mean collision counts and higher mean success rates than AE and VAE.
Reach achieves the fewest collisions, 
while Reach and DeRep achieve the two highest mean success
rates of 37.5\% and 37.7\%, respectively. 

\subsection{Reinforcement Learning Evaluation}
\label{sec:rl_eval}
\begin{figure}[t]
    \centering
    \includegraphics[width=\linewidth]{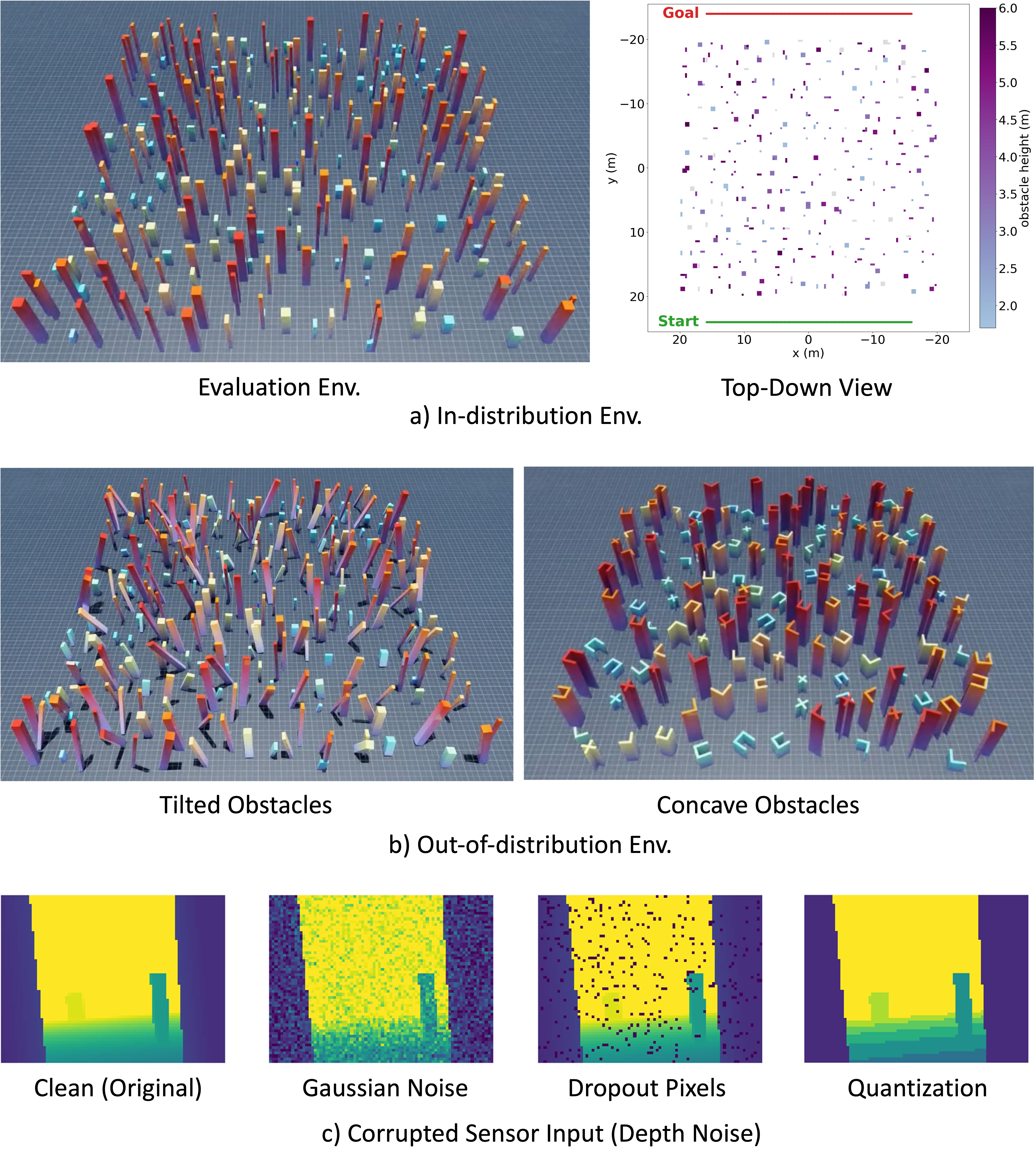}
    \caption{Evaluation conditions: (a) In-distribution (ID). (b) Out-of-distribution (OOD) environments
    with tilted obstacles and concave obstacles. (c) A depth observation before and after
    corruption.}
    \label{fig:eval_conditions}
\end{figure}

\begin{figure}[t]
    \centering
    \includegraphics[width=\linewidth]{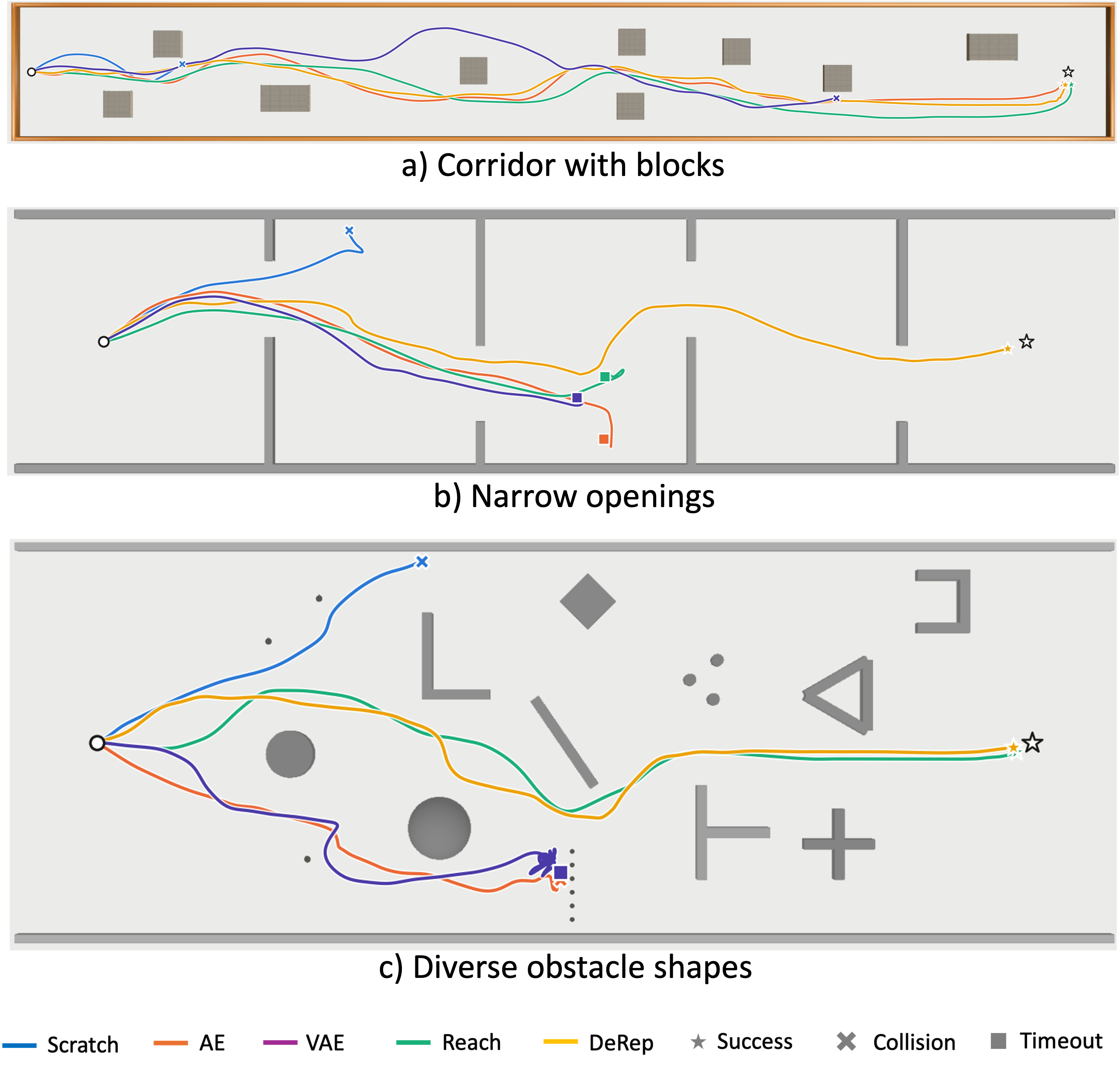}
    \caption{Qualitative comparison in Gazebo. Trajectories of the
PPO policies with each representation in (a) a corridor with box
obstacles, (b) consecutive narrow openings, and (c) diverse obstacle
shapes. All methods share the same start (white circle) and goal
(white star).}
    \label{fig:gazebo_eval}
\end{figure}

\begin{figure*}[t]
    \centering
    \includegraphics[width=\textwidth]{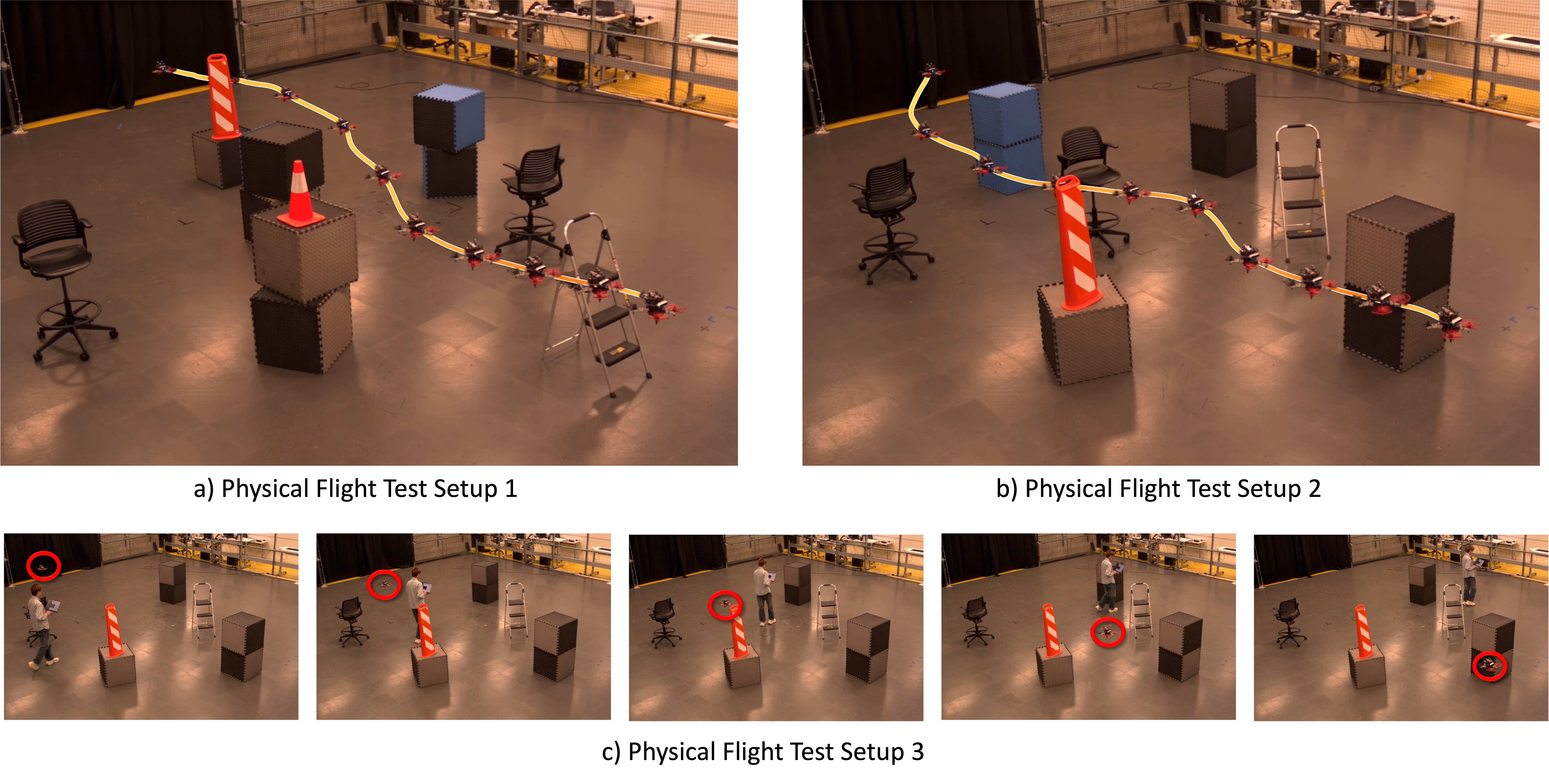}
    \caption{Real-world flight experiments with the DeRep policy.
Snapshots are shown in temporal order from left to right, with the
quadrotor highlighted by red circles. (a)-(b) Two arrangements of
blocks, traffic cones, chairs, and stepladders. (c) A setup
with a person walking through the scene.}
    \label{fig:real_eval}
\end{figure*}

We further evaluate the representations using reinforcement learning in Isaac Sim.
Pretrained encoders remain frozen, while Scratch is trained jointly
with the policy. We use three training seeds per
objective and evaluate the policies under three categories of conditions:
in-distribution environments (ID), unseen obstacle geometry (OOD), and corrupted depth observations (Noise). Example evaluation environments are shown in Fig.~\ref{fig:eval_conditions}. The ID evaluation uses obstacle geometry from the policy training
distribution without added depth noise.
The OOD evaluation includes two settings that are not encountered during policy training: tilted obstacles with roll and pitch angles
within $\pm25^\circ$, and concave obstacles including L-, U-, cross-shaped structures.
The Noise evaluation applies additive Gaussian noise,
per-pixel dropout, and depth quantization to ID depth
observations at three severity levels: additive Gaussian noise with $\sigma = 0.01/0.05/0.10$, random pixel dropout setting $1\%/5\%/10\%$ of the pixels to zero, and uniform quantization to $32/16/8$ depth levels. The same start-goal sampling method is used across all the evaluation conditions.

\textbf{In-distribution performance.}
As shown in Table~\ref{tab:representation_comparison_full}, all the pretrained encoders outperform Scratch. DeRep achieves the highest
mean success rate, exceeding Reach and AE by 15.3 and 16.7 percentage points, respectively. Unlike the imitation learning results,
PPO shows a clear difference in mean performance between the joint objective and either objective alone.
Reach and AE perform similarly in this condition, and both achieve higher mean success rates than VAE and Scratch.

\textbf{Generalization to unseen obstacle geometry.}
All pretrained representations achieve higher mean success
rates than Scratch under both OOD conditions.
Both DeRep and Reach further
outperform AE and VAE in both settings.
DeRep achieves the highest mean success rate under tilted
obstacles, exceeding Reach by 14.0 percentage points.
For concave obstacles, Reach achieves a slightly higher
mean success rate than DeRep, although the difference is
small relative to the variation across training seeds.
The advantage over reconstruction alone is particularly
pronounced for concave obstacles: Reach and DeRep exceed
AE by 25.0 and 21.3 percentage points, respectively.
Concave geometry can require larger detour, making local free-space
connectivity more relevant to decision making.
These results support the benefit of reachability supervision
for generalization to unseen obstacle geometry, as both Reach
and DeRep outperform reconstruction alone under both geometry
shifts.

\textbf{Robustness to depth corruption.}
All pretrained representations achieve higher mean success
rates than Scratch at each corruption severity.
Both DeRep and Reach outperform AE and VAE at all three severity
levels. DeRep achieves the highest mean success rates at
low and medium severity, exceeding AE by 21.3 and 19.0
percentage points, respectively.
At high severity, Reach achieves the highest mean success
rate of 59.4\%, followed by DeRep at 50.1\%, compared with
35.0\% for AE.
The advantage of Reach over AE increases from 3.2 percentage
points at low severity to 11.6 at medium severity and 24.4
at high severity. This suggests that the benefits of
reachability supervision become more apparent as depth
observations are increasingly corrupted.
Although the relative performance of Reach and DeRep varies
with corruption severity, both maintain an advantage over
reconstruction-based objective alone, supporting the value of reachability supervision for robustness to depth corruption.

Overall, DeRep achieves the highest mean success rate in
four of the six PPO evaluation conditions, while Reach
performs best under concave obstacles and high depth
corruption. Both reachability-based objectives outperform
reconstruction alone across all tested PPO conditions.
Together with the linear \rev{policy} probe and imitation learning
results, these findings demonstrate consistent benefits
of incorporating reachability supervision into depth
representation pretraining across different downstream
learning methods and evaluation conditions.


\subsection{Qualitative Comparison in Gazebo}
\label{sec:gazebo_eval}

To complement the quantitative results with a visual comparison of
navigation behavior, we deploy the PPO policies zero-shot in Gazebo. For each representation, we deploy the
policy from the training seed with the highest ID success rate.
We construct three representative environments
(Fig.~\ref{fig:gazebo_eval}): (a) a corridor with box-shaped obstacles,
(b) a sequence of walls with narrow, laterally offset openings that
must be traversed consecutively, and (c) an open area containing
obstacles of diverse shapes not encountered during policy training,
including L-, T-, and cross-shaped structures, spheres, a slanted
bar, and thin cylinders. All methods share the same start and goal
within each environment, and we visualize the resulting trajectory
of each method.

In the corridor environment, AE, Reach, and DeRep all reach the
goal, whereas Scratch and VAE fail due to collision with the obstacles. In the narrow-opening
environment, only DeRep successfully reaches the goal without collision. AE, Reach, and VAE
pass the first two openings but fail to move toward the next offset opening and time out, while Scratch collides with the wall on the side.  In the diverse-obstacle environment, AE and VAE oscillate in front of the row of thin cylinders
until timeout, and Scratch collides with the boundary wall. Reach and DeRep
follow a similar route and successfully reach the goal. 
Overall, DeRep reaches the goal in all three illustrated
environments. These examples complement the quantitative
results in Sec.~\ref{sec:rl_eval} by showing successful navigation in a corridor, through consecutive
narrow openings, and around diverse obstacle shapes.

\subsection{Physical Flight Tests}
\label{sec:real_eval}

We deploy the PPO policy with the frozen DeRep encoder zero-shot on a
quadrotor equipped with a RealSense D435i depth camera and an onboard Jetson Orin NX (Fig.~\ref{fig:cover}). The policy runs onboard at 20~Hz. Robot state estimation is
provided by a motion capture system. Depth images are resized to $60\times80$, clipped at
5~m and normalized as in pretraining.

We test the policy in three indoor setups with different
arrangements of blocks, traffic cones, chairs, and
stepladders (Fig.~\ref{fig:real_eval}).
Setups~1 and 2 use different obstacle arrangements, and Setup~3
additionally includes a person walking through the scene. These experiments demonstrate zero-shot sim-to-real transfer
of the learned navigation policy, which reaches the goal
without collision in the tested real-world environments.
\section{Conclusion}
This work studied how the objectives for learning depth representations
affect downstream navigation performance. We proposed DeRep,
which combines local BEV reachability supervision with depth
reconstruction, and evaluated the frozen representations using
a linear policy probe, flow-matching imitation learning, and PPO.
DeRep achieved the lowest linear \rev{policy} probe error and outperformed reconstruction alone across all six PPO evaluation conditions. In imitation
learning, both reachability-based objectives achieved higher
mean success rates than reconstruction alone, with similar
mean success rates between Reach and DeRep.
These findings demonstrate consistent benefits of incorporating
reachability supervision into representation pretraining
across the tested downstream learning methods and evaluation
conditions.
We further demonstrated zero-shot deployment of the PPO policy
in Gazebo and on a physical quadrotor.
Future work will extend the representation to incorporate
temporal information and evaluate its benefits in dynamic
environments.



\bibliographystyle{IEEEtran}
\bibliography{bibliography}

@INPROCEEDINGS{VitFly,
  author={Bhattacharya, Anish and Rao, Nishanth and Parikh, Dhruv and Kunapuli, Pratik and Wu, Yuwei and Tao, Yuezhan and Matni, Nikolai and Kumar, Vijay},
  booktitle={2025 IEEE International Conference on Robotics and Automation (ICRA)}, 
  title={Vision Transformers for End-to-End Vision-Based Quadrotor Obstacle Avoidance}, 
  year={2025},
  volume={},
  number={},
  pages={1-8},
  doi={10.1109/ICRA55743.2025.11128042}}

@ARTICLE{MAVRL,
  author={Yu, Hang and Wagter, ChristopheDe and de Croon, Guido C. H. E},
  journal={IEEE Robotics and Automation Letters}, 
  title={MAVRL: Learn to Fly in Cluttered Environments With Varying Speed}, 
  year={2025},
  volume={10},
  number={2},
  pages={1441-1448},
  doi={10.1109/LRA.2024.3522778}}

@INPROCEEDINGS{NoMaD,
  author={Sridhar, Ajay and Shah, Dhruv and Glossop, Catherine and Levine, Sergey},
  booktitle={2024 IEEE International Conference on Robotics and Automation (ICRA)}, 
  title={NoMaD: Goal Masked Diffusion Policies for Navigation and Exploration}, 
  year={2024},
  volume={},
  number={},
  pages={63-70},
  doi={10.1109/ICRA57147.2024.10610665}}

@INPROCEEDINGS{FlowNav,
  author={Gode, Samiran and Nayak, Abhijeet and Oliveira, Débora N.P. and Krawez, Michael and Schmid, Cordelia and Burgard, Wolfram},
  booktitle={2025 IEEE/RSJ International Conference on Intelligent Robots and Systems (IROS)}, 
  title={FlowNav: Combining Flow Matching and Depth Priors for Efficient Navigation}, 
  year={2025},
  volume={},
  number={},
  pages={17762-17768},
  doi={10.1109/IROS60139.2025.11246634}}

@INPROCEEDINGS{CollisionEnc,
  author={Kulkarni, Mihir and Alexis, Kostas},
  booktitle={2024 IEEE International Conference on Robotics and Automation (ICRA)}, 
  title={Reinforcement Learning for Collision-free Flight Exploiting Deep Collision Encoding}, 
  year={2024},
  volume={},
  number={},
  pages={15781-15788},
  doi={10.1109/ICRA57147.2024.10610287}}

@INPROCEEDINGS{CollisionPrediction,
  author={Kulkarni, Mihir and Nguyen, Huan and Alexis, Kostas},
  booktitle={2023 IEEE/RSJ International Conference on Intelligent Robots and Systems (IROS)}, 
  title={Semantically-Enhanced Deep Collision Prediction for Autonomous Navigation Using Aerial Robots}, 
  year={2023},
  volume={},
  number={},
  pages={3056-3063},
  doi={10.1109/IROS55552.2023.10342297}}

@INPROCEEDINGS{NavRep,
  author={Dugas, Daniel and Nieto, Juan and Siegwart, Roland and Chung, Jen Jen},
  booktitle={2021 IEEE International Conference on Robotics and Automation (ICRA)}, 
  title={NavRep: Unsupervised Representations for Reinforcement Learning of Robot Navigation in Dynamic Human Environments}, 
  year={2021},
  volume={},
  number={},
  pages={7829-7835},
  doi={10.1109/ICRA48506.2021.9560951}}

@INPROCEEDINGS{NavDream,
  author={Dugas, Daniel and Andersson, Olov and Siegwart, Roland and Chung, Jen Jen},
  booktitle={2022 IEEE/RSJ International Conference on Intelligent Robots and Systems (IROS)}, 
  title={NavDreams: Towards Camera-Only RL Navigation Among Humans}, 
  year={2022},
  volume={},
  number={},
  pages={2504-2511},
  doi={10.1109/IROS47612.2022.9982045}}

@INPROCEEDINGS{Ego2Map,
  author={Hong, Yicong and Zhou, Yang and Zhang, Ruiyi and Dernoncourt, Franck and Bui, Trung and Gould, Stephen and Tan, Hao},
  booktitle={2023 IEEE/CVF International Conference on Computer Vision (ICCV)}, 
  title={Learning Navigational Visual Representations with Semantic Map Supervision}, 
  year={2023},
  volume={},
  number={},
  pages={3032-3044},
  doi={10.1109/ICCV51070.2023.00284}}

@inproceedings{Mole,
 author = {Bono, Guillaume and Antsfeld, Leonid and Sadek, Assem and Monaci, Gianluca and Wolf, Christian},
 booktitle = {International Conference on Learning Representations},
 editor = {B. Kim and Y. Yue and S. Chaudhuri and K. Fragkiadaki and M. Khan and Y. Sun},
 pages = {31674--31692},
 title = {Learning with a Mole: Transferable latent spatial representations for navigation without reconstruction},
 url = {https://proceedings.iclr.cc/paper_files/paper/2024/file/8667f264f88c7938a73a53ab01eb1327-Paper-Conference.pdf},
 volume = {2024},
 year = {2024}
}

@misc{DeFM,
      title={DeFM: Learning Foundation Representations from Depth for Robotics}, 
      author={Manthan Patel and Jonas Frey and Mayank Mittal and Fan Yang and Alexander Hansson and Amir Bar and Cesar Cadena and Marco Hutter},
      year={2026},
      eprint={2601.18923},
      archivePrefix={arXiv},
      primaryClass={cs.RO},
      url={https://arxiv.org/abs/2601.18923}, 
}

@misc{dinov2,
      title={DINOv2: Learning Robust Visual Features without Supervision}, 
      author={Maxime Oquab and Timothée Darcet and Théo Moutakanni and Huy Vo and Marc Szafraniec and Vasil Khalidov and Pierre Fernandez and Daniel Haziza and Francisco Massa and Alaaeldin El-Nouby and Mahmoud Assran and Nicolas Ballas and Wojciech Galuba and Russell Howes and Po-Yao Huang and Shang-Wen Li and Ishan Misra and Michael Rabbat and Vasu Sharma and Gabriel Synnaeve and Hu Xu and Hervé Jegou and Julien Mairal and Patrick Labatut and Armand Joulin and Piotr Bojanowski},
      year={2024},
      eprint={2304.07193},
      archivePrefix={arXiv},
      primaryClass={cs.CV},
      url={https://arxiv.org/abs/2304.07193}, 
}

@inproceedings{VisualControl,
 author = {Dong, Jiahua and Man, Yunze and Tokmakov, Pavel and Wang, Yu-Xiong},
 booktitle = {International Conference on Learning Representations},
 editor = {C. Vondrick and B. Hariharan and C. Raffel and L. Pinto and D. Yang and A. Faust},
 pages = {35435--35462},
 title = {Capturing Visual Environment Structure Correlates with Control Performance},
 url = {https://proceedings.iclr.cc/paper_files/paper/2026/file/3bca7054c944d9d254adaa953307bdd8-Paper-Conference.pdf},
 volume = {2026},
 year = {2026}
}

@misc{R3M,
      title={R3M: A Universal Visual Representation for Robot Manipulation}, 
      author={Suraj Nair and Aravind Rajeswaran and Vikash Kumar and Chelsea Finn and Abhinav Gupta},
      year={2022},
      eprint={2203.12601},
      archivePrefix={arXiv},
      primaryClass={cs.RO},
      url={https://arxiv.org/abs/2203.12601}, 
}

@article{LongRange,
author = {Fan Yang and Per Frivik and David Hoeller and Chen Wang and Cesar Cadena and Marco Hutter},
title ={Spatially-enhanced recurrent memory for long-range mapless navigation via end-to-end reinforcement learning},

journal = {The International Journal of Robotics Research},
volume = {0},
number = {0},
pages = {02783649251401926},
year = {2025},
doi = {10.1177/02783649251401926},
URL = { https://doi.org/10.1177/02783649251401926},
}

@ARTICLE{HighSpeedRL,
  author={Zhang, Jiarui and Lei, Chengyong and Dai, Chengjiang and Hoi, Kenghou and Wang, Lijie and Han, Zhichao and Gao, Fei},
  journal={IEEE Robotics and Automation Letters}, 
  title={High-Speed Vision-Based Flight in Clutter With Safety-Shielded Reinforcement Learning}, 
  year={2026},
  volume={11},
  number={8},
  pages={9559-9566},
  doi={10.1109/LRA.2026.3703580},
  ISSN={2377-3766},
  month={Aug},}

@InProceedings{SimCLR,
  title = 	 {A Simple Framework for Contrastive Learning of Visual Representations},
  author =       {Chen, Ting and Kornblith, Simon and Norouzi, Mohammad and Hinton, Geoffrey},
  booktitle = 	 {Proceedings of the 37th International Conference on Machine Learning},
  pages = 	 {1597--1607},
  year = 	 {2020},
  editor = 	 {III, Hal Daumé and Singh, Aarti},
  volume = 	 {119},
  series = 	 {Proceedings of Machine Learning Research},
  month = 	 {13--18 Jul},
  publisher =    {PMLR},
  url = 	 {https://proceedings.mlr.press/v119/chen20j.html},
}

@InProceedings{pmlr-v202-hu23h,
  title = 	 {For Pre-Trained Vision Models in Motor Control, Not All Policy Learning Methods are Created Equal},
  author =       {Hu, Yingdong and Wang, Renhao and Li, Li Erran and Gao, Yang},
  booktitle = 	 {Proceedings of the 40th International Conference on Machine Learning},
  pages = 	 {13628--13651},
  year = 	 {2023},
  editor = 	 {Krause, Andreas and Brunskill, Emma and Cho, Kyunghyun and Engelhardt, Barbara and Sabato, Sivan and Scarlett, Jonathan},
  volume = 	 {202},
  series = 	 {Proceedings of Machine Learning Research},
  month = 	 {23--29 Jul},
  publisher =    {PMLR},
  url = 	 {https://proceedings.mlr.press/v202/hu23h.html}
}

@INPROCEEDINGS{kendall2018,
  author={Cipolla, Roberto and Gal, Yarin and Kendall, Alex},
  booktitle={2018 IEEE/CVF Conference on Computer Vision and Pattern Recognition}, 
  title={Multi-task Learning Using Uncertainty to Weigh Losses for Scene Geometry and Semantics}, 
  year={2018},
  volume={},
  number={},
  pages={7482-7491},
  doi={10.1109/CVPR.2018.00781}}

@misc{xia2018gibson,
      title={Gibson Env: Real-World Perception for Embodied Agents}, 
      author={Fei Xia and Amir Zamir and Zhi-Yang He and Alexander Sax and Jitendra Malik and Silvio Savarese},
      year={2018},
      eprint={1808.10654},
      archivePrefix={arXiv},
      primaryClass={cs.AI},
      url={https://arxiv.org/abs/1808.10654}, 
}

@InProceedings{savva2019habitat,
author = {Savva, Manolis and Kadian, Abhishek and Maksymets, Oleksandr and Zhao, Yili and Wijmans, Erik and Jain, Bhavana and Straub, Julian and Liu, Jia and Koltun, Vladlen and Malik, Jitendra and Parikh, Devi and Batra, Dhruv},
title = {Habitat: A Platform for Embodied AI Research},
booktitle = {Proceedings of the IEEE/CVF International Conference on Computer Vision (ICCV)},
month = {October},
year = {2019}
}

@misc{STERLING,
      title={STERLING: Self-Supervised Terrain Representation Learning from Unconstrained Robot Experience}, 
      author={Haresh Karnan and Elvin Yang and Daniel Farkash and Garrett Warnell and Joydeep Biswas and Peter Stone},
      year={2023},
      eprint={2309.15302},
      archivePrefix={arXiv},
      primaryClass={cs.RO},
      url={https://arxiv.org/abs/2309.15302}, 
}

@inproceedings{OVRL,
title={Offline Visual Representation Learning for Embodied Navigation},
author={Karmesh Yadav and Ram Ramrakhya and Arjun Majumdar and Vincent-Pierre Berges and Sachit Kuhar and Dhruv Batra and Alexei Baevski and Oleksandr Maksymets},
booktitle={Workshop on Reincarnating Reinforcement Learning at ICLR 2023},
year={2023},
url={https://openreview.net/forum?id=Spfbts_vNY}
}

@article{schulman2017ppo,
  author       = {John Schulman and
                  Filip Wolski and
                  Prafulla Dhariwal and
                  Alec Radford and
                  Oleg Klimov},
  title        = {Proximal Policy Optimization Algorithms},
  journal      = {CoRR},
  volume       = {abs/1707.06347},
  year         = {2017},
  url          = {http://arxiv.org/abs/1707.06347},
  eprinttype   = {arXiv},
  eprint       = {1707.06347},
  bibsource    = {dblp computer science bibliography, https://dblp.org}
}

@misc{ASIP,
      title={ASIP-Planner: Adaptive Planning for UAV Surface Inspection in Partially Known Indoor Environments}, 
      author={Hanyu Jin and Zhefan Xu and Haoyu Shen and Xinming Han and Kanlong Ye and Kenji Shimada},
      year={2026},
      eprint={2605.11119},
      archivePrefix={arXiv},
      primaryClass={cs.RO},
      url={https://arxiv.org/abs/2605.11119}, 
}

@ARTICLE{FUEL,
  author={Zhou, Boyu and Zhang, Yichen and Chen, Xinyi and Shen, Shaojie},
  journal={IEEE Robotics and Automation Letters}, 
  title={FUEL: Fast UAV Exploration Using Incremental Frontier Structure and Hierarchical Planning}, 
  year={2021},
  volume={6},
  number={2},
  pages={779-786},
  doi={10.1109/LRA.2021.3051563}}

@ARTICLE{NavRL,
  author={Xu, Zhefan and Han, Xinming and Shen, Haoyu and Jin, Hanyu and Shimada, Kenji},
  journal={IEEE Robotics and Automation Letters}, 
  title={NavRL: Learning Safe Flight in Dynamic Environments}, 
  year={2025},
  volume={10},
  number={4},
  pages={3668-3675},
  doi={10.1109/LRA.2025.3546069}}

@ARTICLE{FastPlanner,
  author={Zhou, Boyu and Gao, Fei and Wang, Luqi and Liu, Chuhao and Shen, Shaojie},
  journal={IEEE Robotics and Automation Letters}, 
  title={Robust and Efficient Quadrotor Trajectory Generation for Fast Autonomous Flight}, 
  year={2019},
  volume={4},
  number={4},
  pages={3529-3536},
  doi={10.1109/LRA.2019.2927938}}

@ARTICLE{EgoPlanner,
  author={Zhou, Xin and Wang, Zhepei and Ye, Hongkai and Xu, Chao and Gao, Fei},
  journal={IEEE Robotics and Automation Letters}, 
  title={EGO-Planner: An ESDF-Free Gradient-Based Local Planner for Quadrotors}, 
  year={2021},
  volume={6},
  number={2},
  pages={478-485},
  doi={10.1109/LRA.2020.3047728}}

\end{document}